\documentclass[10pt,twocolumn,letterpaper,amsart]{article}

\usepackage[pagenumbers]{cvpr} 

\usepackage{graphicx}
\usepackage{amsmath}
\usepackage{amssymb}
\usepackage{booktabs}
\usepackage{enumitem} 
\usepackage{xcolor}
\usepackage{tabularx}
\usepackage{multicol, blindtext}

\usepackage[pagebackref,breaklinks,colorlinks]{hyperref}

\usepackage[capitalize]{cleveref}
\crefname{section}{Sec.}{Secs.}
\Crefname{section}{Section}{Sections}
\Crefname{table}{Table}{Tables}
\crefname{table}{Tab.}{Tabs.}

\def\cvprPaperID{*****} 
\def\confName{CVPR}
\def\confYear{2023}

\begin{document}

\title{ MultiEarth 2023 Deforestation Challenge - Team FOREVER }

\author{Seunghan Park\thanks{Equal contribution} \quad Dongoo Lee \thanks{Equal contribution} \quad Yeonju Choi \thanks{Equal contribution} \quad  SungTae Moon\footnotemark[1]  \thanks{Corresponding author}\\
 Korea Aerospace Research Institute, Korea University of Technology and Education \\
{ \tt\small \ pcappli@koreatech.ac.kr,ldg810@kari.re.kr,choiyj@kari.re.kr,stmoon@koreatech.ac.kr}}
\maketitle

\begin{abstract}
It is important problem to accurately estimate deforestation of satellite imagery since this approach can analyse extensive area without direct human access. However, it is not simple problem because of difficulty in observing the clear ground surface due to extensive cloud cover during long rainy season. In this paper, we present a multi-view learning strategy to predict deforestation status in the Amazon rainforest area with latest deep neural network models. Multi-modal dataset consists of three types of different satellites imagery, Sentinel-1, Sentinel-2 and Landsat 8 is utilized to train and predict deforestation status. MMsegmentation framework is selected to apply comprehensive data augmentation and diverse networks. The proposed method effectively and accurately predicts the deforestation status of new queries.
\end{abstract}

\section{Introduction}
\label{sec:intro}
The Amazon is the most important and priceless resource on Earth. It plays a key role in reducing the negative consequences of climate change by actively absorbing greenhouse gases and producing oxygen, acting as a crucial regulator. However, the problem of Amazon deforestation is becoming increasingly serious. Unregulated deforestation has negative effects, such as the eradication of ecosystems, loss of biodiversity, soil erosion, and accelerated global warming. A daily average of 2,300 hectares of forest were destroyed in 2020 \cite{MapBiomas} and as a result, reducing the rate of Amazonian deforestation through the creation of protected zones has gained international attention. Effective forest management, including strengthened forest monitoring, is essential to addressing this urgent issue\cite{banskota2014forest}.

Recently, as the types of available satellite images have been diversified, the spatial resolution of satellites is improving, and deep learning is showing remarkable performance in the image analysis area \cite{de2022deforestation, john2022attention}. Especially, the multi-sensor data fusion strategy combining SAR ( Synthetic Aperture Radar) and optical sensors such as Landsat has improved the accuracy of forest mapping \cite{walker2010large}. The MultiEarth 2023 is the second CVPR workshop utilizing multi-satellites imagery to help monitor and analyze the health of these Earth ecosystems \cite{cha2022multiearth, cha2023multiearth}. The MultiEarth 2023 Challenge's objective is to combine optical and SAR imagery to carry out continuous assessments of forest monitoring at any time and in all weather. This research is the extending work of last year's deforestation challenge study \cite{lee2022multiearth}

The main contributions of this research are summarized as follows.
\begin{itemize}
\item Post-processing with effective cloud removal to accurately predict deforestation with multi-modal dataset is proposed.
\item Time-series analysis method with adjacent month data is proposed.
\end{itemize}

The remainder of this paper is organized as follows. \cref{sec:data} introduced the contents and generation method of the dataset in detail and \cref{sec:metho} described the proposed methodology for deforestation detection and in \cref{sec:res}, prediction results are presented. The conclusion is presented in \cref{sec:con}.

\section{Dataset}
\label{sec:data}
\begin{figure*}
  \centering
  \includegraphics[height=70mm, width=130mm]{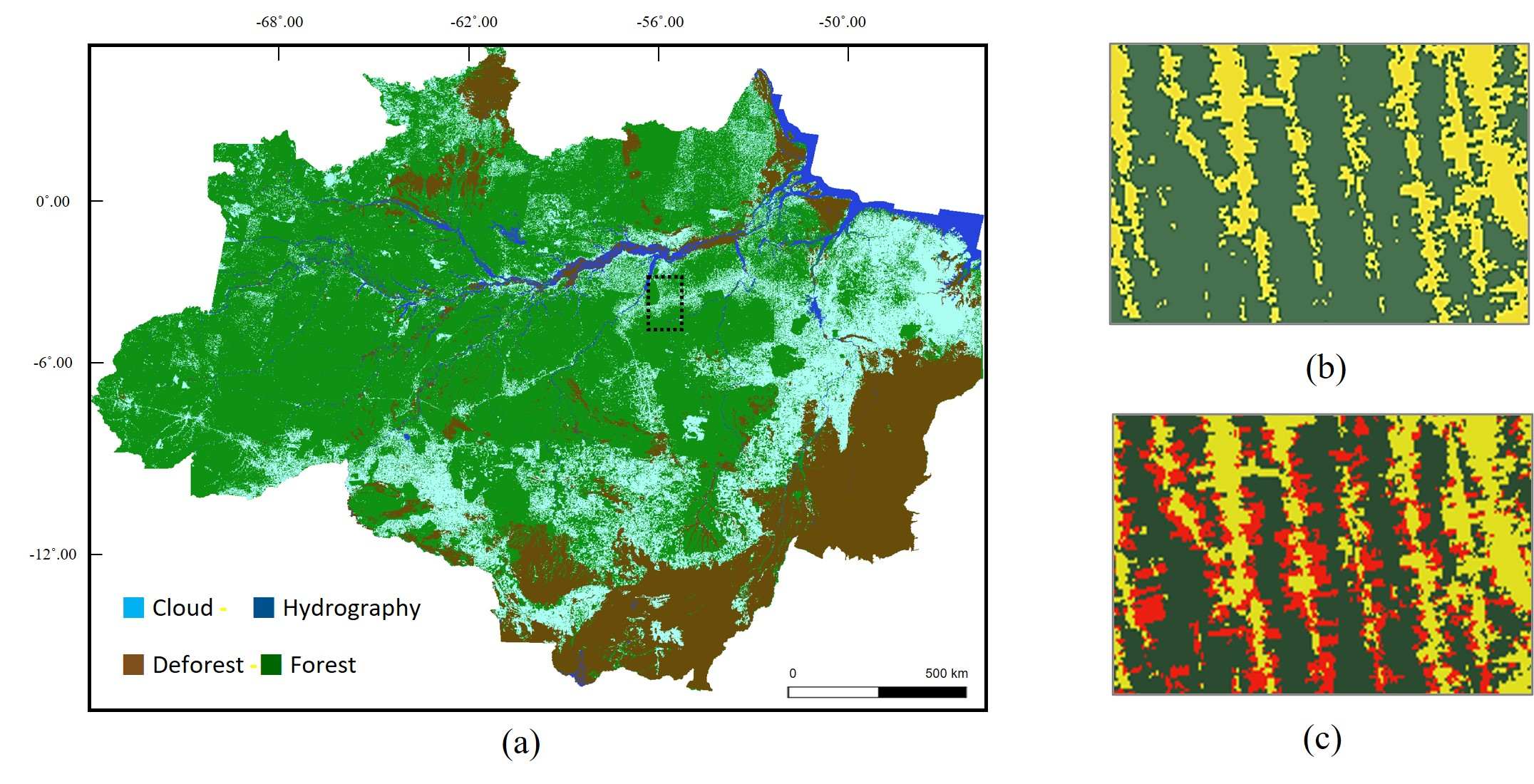}
  \caption{(a) Deforestation map of Brazil, (b) deforestation in test region until 2007 highlighted in yellow, and (c) deforestation in test region since 2007 highlighted in red.}
  \label{fig:roi}
\end{figure*}

The area of interest in this study is the area bounded by (latitude (LAT) :-3.33$^{\circ}$ $\sim$ -4.39$^{\circ}$ , longitude (LON) :-54.48$^{\circ}$ $\sim$ 55.2$^{\circ}$), which is one of the areas where deforestation occurs very frequently in the Amazon. The region comprises a portion of dense tropical Amazon rainforest in Para, Brazil, containing thousands of species of broad-leaved evergreen trees. Historically, this has been one of the areas with the highest tree loss rate in the Amazon region, with pastures in the region nearly doubling between 1997 and 2007. As shown in Figure \cref{fig:roi} (b) from PRODES (Brazilian Amazon Rainforest Monitoring Program by Satellite) and the DETER (Real-time Deforestation Detection System), it can be seen that deforestation in the area of interest has expanded before and after 2007 \cite{fg2019terrabrasilis}. 

We used four datasets obtained from Sentinel-1 (Sen1), Sentinel-2 (Sen2), Landsat 5 (Land5), and Landsat 8 (Land8). The four types of satellites are equipped with different sensors, and have different temporal/spatial resolutions and acquisition cycles. Specific specifications for the utilization dataset are described in \cref{tab:seldata}.

\begin{table}
\centering
\begin{tabular}{ccccc}
\toprule 
\textbf{Satellite} & \textbf{Time Range} & \textbf{ Bands} & \textbf{Resolution} \\
\midrule 
\textbf{Landsat 5}        & 1984-2012         & 8           & 30 m   \\
\textbf{Landsat 8}        & 2013-2021         & 9           & 30 m   \\
\textbf{Sentinel-1}      & 2014-2021         & 2           & 10 m   \\
\textbf{Sentinel-2}      & 2018-2021         & 12          & 10 m   \\
\textbf{Labeling }       & 2016-2021          & 1          & 10 m    \\
\bottomrule
\end{tabular}
\caption{ Satellites and data specification.}
\label{tab:seldata}
\end{table}

The final size of all dataset images is 256 x 256, and in the labeling image, the deforestation pixels are set to 1 and the forest or background pixels are set to 0.

\section{Methodology}
\label{sec:metho}
\begin{figure}[h]
  \centering
  \includegraphics[width=80mm,height=45mm]{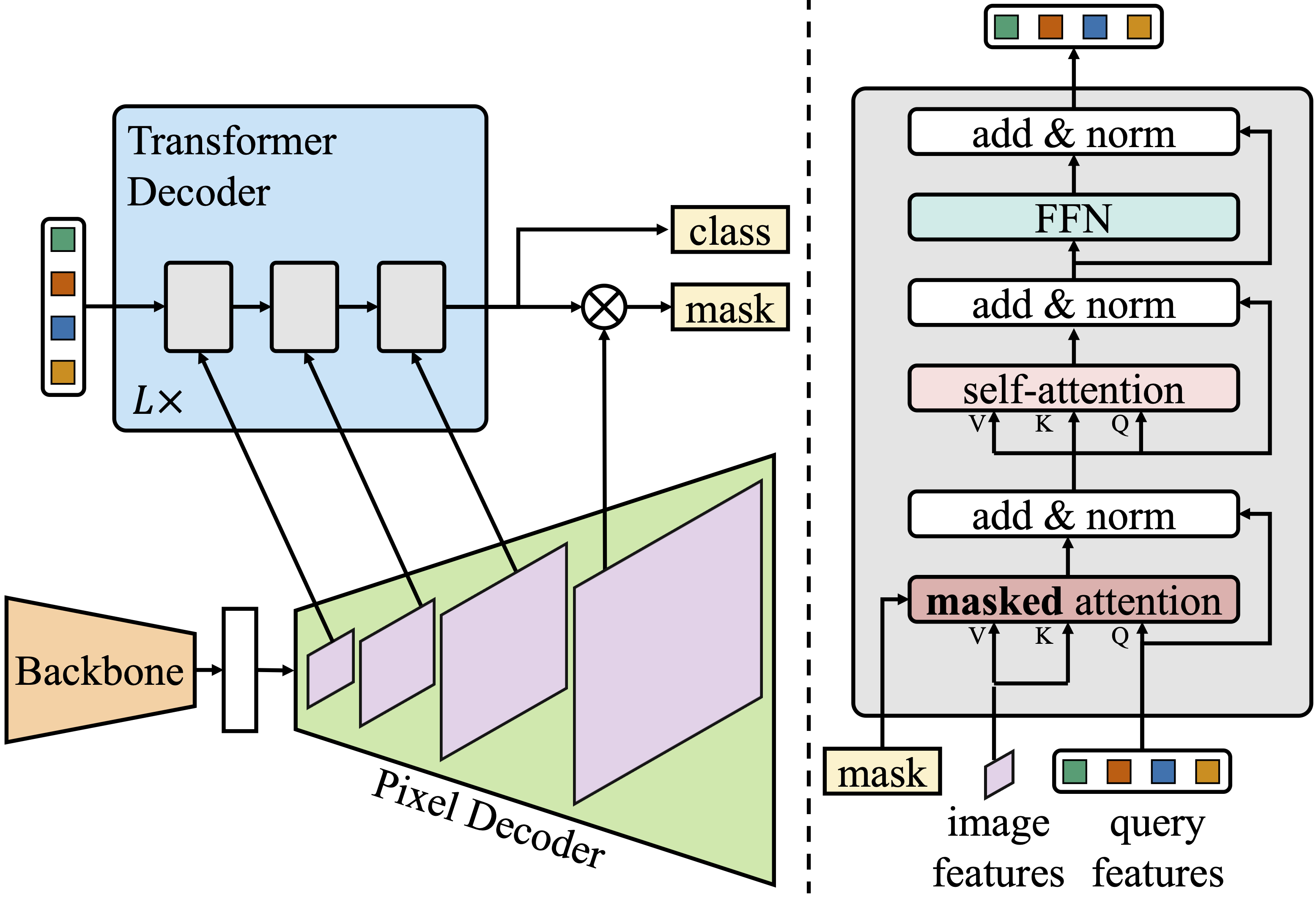}
  \caption{Mask2Former network architecture \cite{mask2former}.}
  \label{fig:mask2former}
\end{figure}

The deforestation estimation challenge of MultiEarth is to determine whether a region is deforested or not. To solve the problem, we adopted Masked-attention Mask Transformer (Mask2Former)\cite{mask2former}, a new architecture capable of addressing any image segmentation task as depicted in \cref{fig:mask2former}. With the Mask2Former deep neural network, we focused on data pre-processing and post-processing to improve performance. The overall procedures for deforestation estimation method is as follows.

\subsection{Pre-processing}
\label{sec:preprocessing}
To use diverse data augmentation libraries, which are mostly support only 3-channel image, we only select RGB bands of Sen2 and Land8 imagery data. To make 3-channel image, a mock band filled with 0 is inserted to Sen1 imagery data. The pre-processing procedures of training dataset are summarized as follows.
\begin{enumerate}[label=\roman*]
\item RGB bands for Sen2 and Land8 are selected and VV/VH bands are selected for Sen1.
\item Generate a training input image with selected bands. a mock band is inserted to Sen1 to make it 3 channel image. The shape of the processed images are (3, 256, 256) for all types of satellite.
\item Remove lowest 2\% and highest 2\% data in the image and normalized to [0,1].
\end{enumerate}

\subsection{Training Network}
\label{sec:trainnet}

To detect the deforested area in this challenge, we tried to adopt several networks using MMSegmentation \cite{mmseg2020} which is an open source semantic segmentation toolbox based on PyTorch. Through multiple experiments, we have discovered that Mask2Former \cite{mask2former} is the most suitable network for this challenge. In addition, there were differences in pixel accuracy depending on the backbone network of the mask2former network, as shown in \cref{tab:perf_backbone}. Therefore, we selected the best backbone network for each satellite.

\begin{table}[h]
\centering 
\begin{tabular}{p{2.5cm} p{2cm} p{2cm}}
\toprule
\textbf{Method} & \textbf{ResNet-50} & \textbf{Swin-L}   \\ 
\midrule
Landsat 8 & 79.445    & 80.84   \\ 
Sentinel-1  & 82.13    & 79.42   \\ 
Sentinel-2  & 78.145    & 77.41   \\ 
\bottomrule
\end{tabular}
\caption{Pixel accuracy comparison according to backbone network}
\label{tab:perf_backbone}
\end{table}

In this challenge, there are two detection classes. The deforested area, which is the target, is labeled as class 1, while the forested/other area is labeled as class 0. To compensate for the class imbalance, we adopted the combination of  Binary Cross-Entropy loss\cite{yi2004automated} and the Dice loss\cite{sudre2017generalised}. In addition, the model training is performed with AdamW optimization, and the learning rate is adjusted by checking for validation loss every two epochs. 

\subsection{Post-processing}
\label{sec:post}

\begin{figure*}[ht]
\centering
\includegraphics[width=170mm]{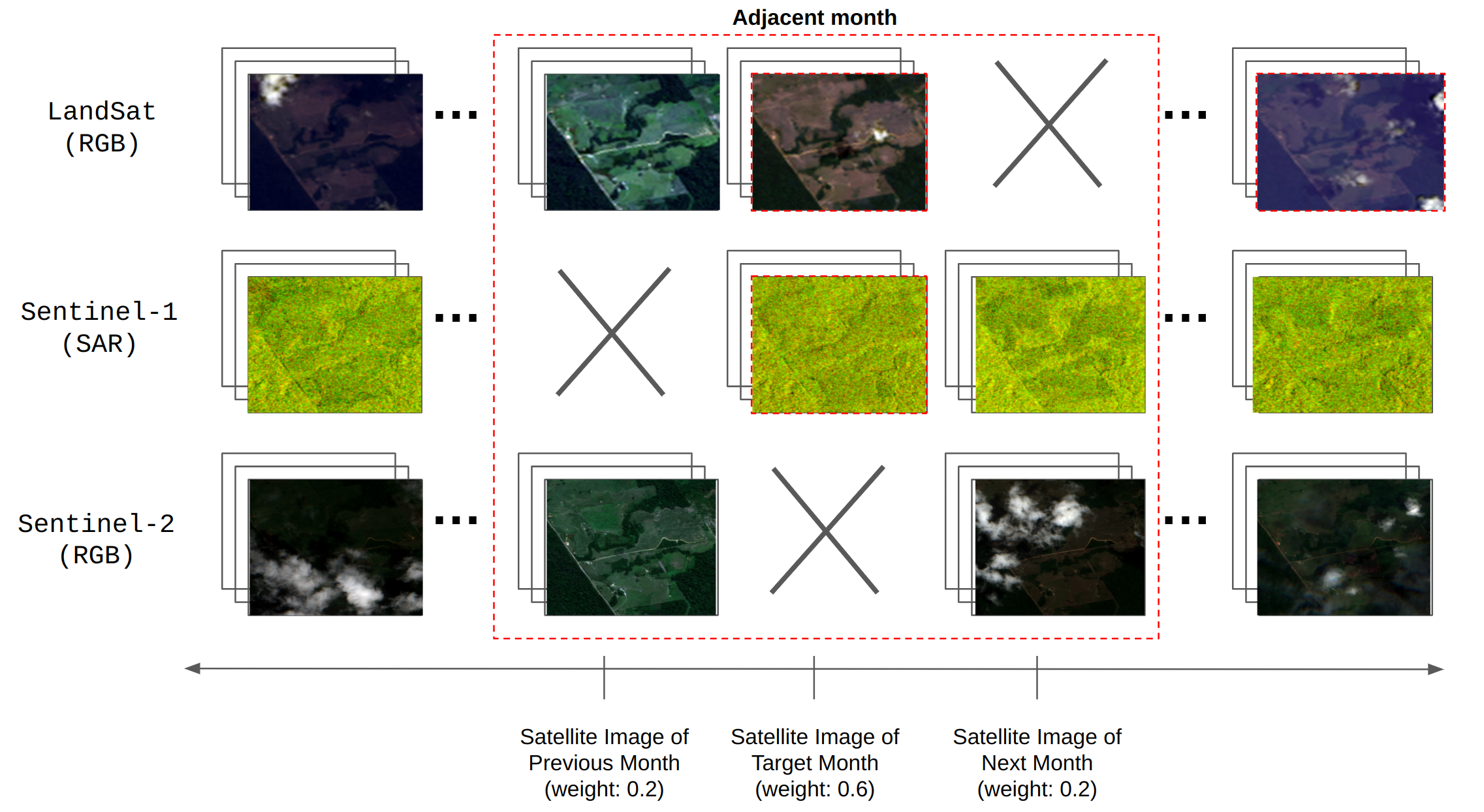}
\caption{Post processing for the adjacent month.}
\label{fig:month_imgs}
\end{figure*}

The deforestation estimation test queries consist of satellite images captured at specific times and locations, expressed in latitude and longitude coordinates. However, there are many cases where there are no satellite images available at the given query, and in some cases, there are only one or two images available as shown in \cref{fig:month_imgs}. In addition, test images data of the same region at different times are supported. Therefore, to address the absence of image data or insufficient information at the given query, we leveraged time-series data that incorporated images captured at the same location during similar periods.For this post processing, the deforestation estimation output of image from the adjacent month for a specific query are generated using different network according to Sen1, Sen2, and Land8. To enhance the importance of the images captured in the current month, higher weights were assigned compared to the preceding and succeeding months.

\begin{figure*}[ht]
\centering
\includegraphics[width=170mm]{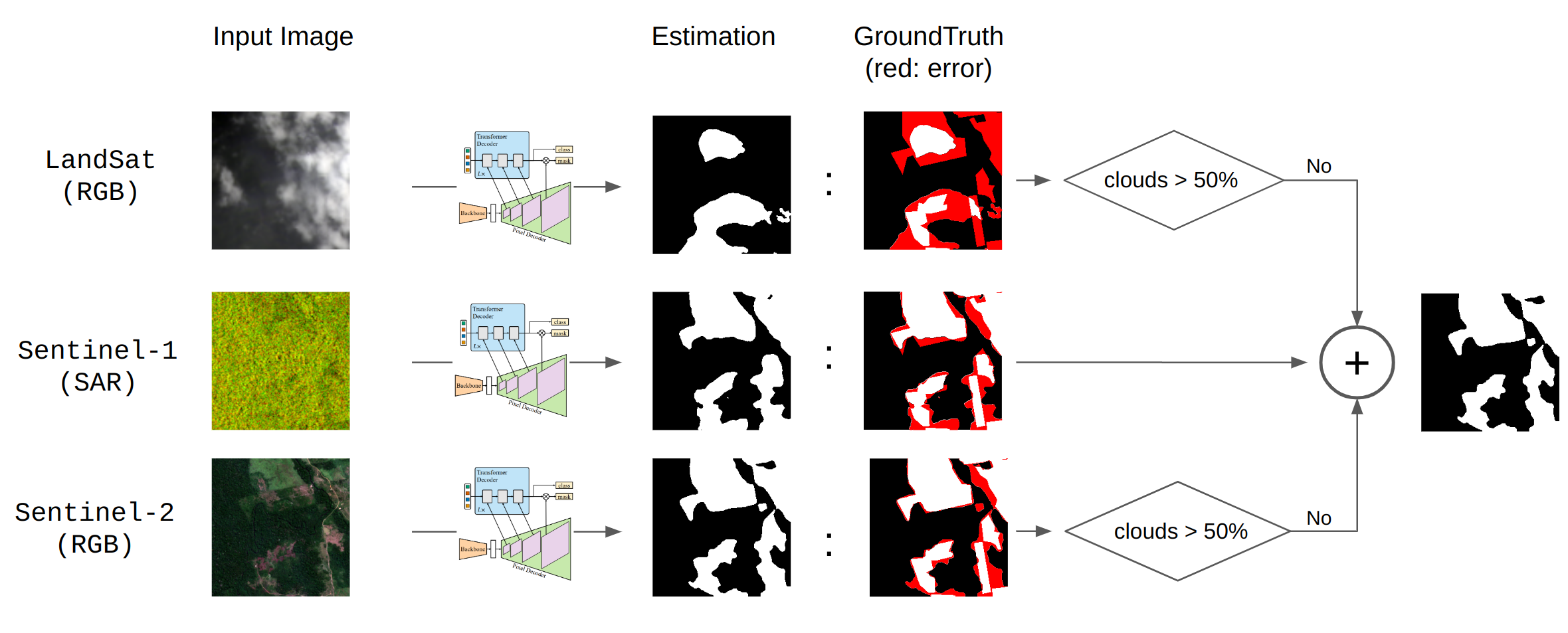}
\caption{Post processing for cloud removal.}
\label{fig:cloud_removal}
\end{figure*}

On the other hands, the clouds in the image can lead to misleading results when estimating deforestation areas. To enhance the deforestation estimation performance, we exclude images that contain a significant amount of clouds. Especially, it was observed that the accuracy of the estimation decreased for Land8 and Sen2 images with a significant presence of clouds, as shown in \cref{fig:cloud_removal}. This was attributed to the selection of RGB bands unlike Sen1. Therefore, for Land8 and Sen2, which are affected by clouds, images with a significant cloud presence were removed. The removal criterion was determined by a simple rule: images in which all RGB values exceeded 160 were classified as clouds, with the threshold set at over 50\% of the total image pixel. In case of Sen1 images, no cloud removal was performed because it ware acquired using SAR. 

\begin{figure*}[ht]
\centering
\includegraphics[width=170mm]{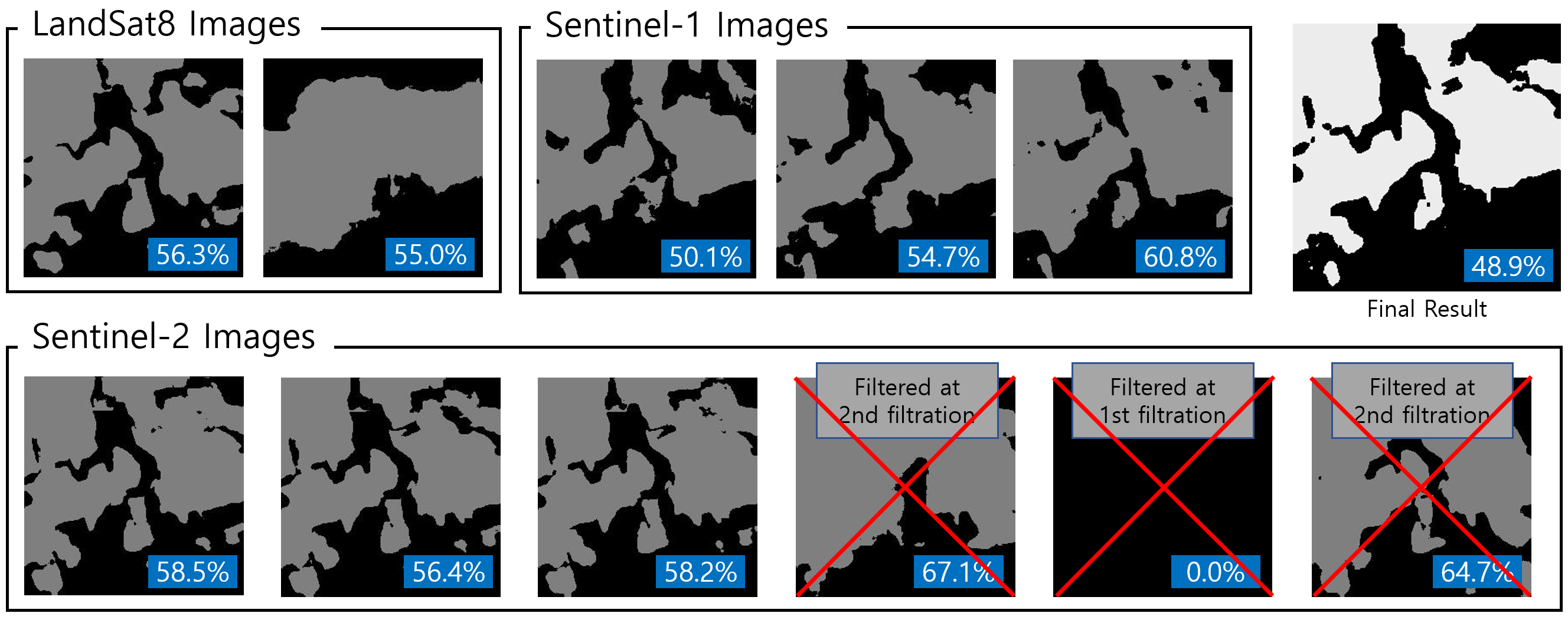}
\caption{Post processing for outlier filter}
\label{fig:outlier_filter}
\end{figure*}

Despite learning from the same network, some of the prediction results were completely black (no deforestation), and some present plausible deforestation results. Therefore, two-step filtering was applied to the output result to ensure the overall detection performance as shown in \cref{fig:outlier_filter}. The clear outliers were filtered without three-sigma range ($\mu$ $\pm$ 3$\sigma$) at the first filtration step. Second filtration used one-sigma range  ($\mu$ $\pm$ $\sigma$) using predicted deforestation percentage in the images. Here, $\mu$ is the mean and $\sigma$ is the variation of deforestation ratio for the predicted output.

Using the filtered output images from the previous step, all images were averaged to create a single binary image. In this image, each pixel was assigned a value of 1 if the probability of deforestation in that pixel exceeded a specific threshold (we set the threshold at 40\%). Conversely, if the probability was below the threshold, the pixel value was assigned as 0.

Finally, the noise in binary image is removed  with the morphology opening operation $\ominus$ as described in \cref{eq:opening}  that is dilation operation of erosion result.

\begin{equation}
    I\circ M  = (I \ominus M)\oplus M,
     \label{eq:opening}
\end {equation}
where, \textit{I}, \textit{M} are the original image and a structuring element and $\ominus$, $\oplus$ are the erosion and dilation operation respectively. The erosion  operation eliminates small objects and dilation restores the size and shape of the remaining objects in image.

\section{Results}
\label{sec:res}

The test set comprises 1,000 queries spanning from August 2016 to August 2021, covering 135 regions. The amount of image information provided for each query was inconsistent. Therefore, we focus on the post processing to improve the pixel accuracy. To evaluate the post-processing performance, each method is assessed from a evaluation website. Especially, With the cloud removal post processing, the pixel accuracy is increased up to 90.546. In addition, when there are only a few satellite images available at the given query time, considering time-series data increases pixel accuracy.

With cloud removal and time adjacent image data in the post-processing procedure as described in \cref{sec:post}, final detection performance has been improved from the initial results. As a result, the proposed method finally achieves 91.13 pixel accuracy, 0.88 F1-score and 0.81 IoU for the test set as represented in \cref{tab:result}.

\begin{table*}[hbt!]
\centering
\begin{tabular}{p{8cm} p{2.5cm} p{2.5cm} p{2.5cm}}
\toprule
\textbf{Method} & \textbf{Pixel Accuracy} & \textbf{F1 Score} & \textbf{IoU}    \\ 
\midrule
Mask2Former (original) & 89.762        & 0.865   & 0.784 \\ 
Mask2Former (with cloud removal) & 90.536        & 0.879   & 0.801 \\ 
Mask2Former (with cloud removal and adjacent month) & 91.139        & 0.888   & 0.814 \\ 
\bottomrule
\end{tabular}
\caption{Final statistics for test set evaluation.}
\label{tab:result}
\end{table*}

\section{Conclusion}
\label{sec:con}

In the MultiEarth 2023 deforestation estimation challenge, three types of satellite imagery (Sentinel-1, Sentinel-2, and Landsat 8) are provided as a multi-modal dataset. The MMsegmentation framework is used to adopt the latest deep neural networks and diverse augmentation libraries. Since data augmentation libraries mostly support three-channel images, we select the RGB bands of satellite images. Two different backbones for the Mask2Former method are compared, and the better backbones are selected for each satellite data. Finally, post-processing is performed to remove cloud-covered images, filter outlier prediction results, average them with previous and next month data, and denoising operations. The proposed method achieves the best scores in all evaluation metrics (pixel accuracy, F1, and IoU).

{\small
\bibliographystyle{ieee_fullname}
\bibliography{egbib}
}

\end{document}